# Knowledge Extraction and Distillation from Large-Scale Image-Text Colonoscopy Records Leveraging Large Language and Vision Models

Shuo Wang[1,2,3,4&*], Yan Zhu[4,5&], Xiaoyuan Luo[1,2&], Zhiwei Yang[1,2,6&], Yizhe Zhang[7], Peiyao Fu[4,5], Manning Wang[1,2], Zhijian Song[1,2], Quanlin Li[4,5*], Pinghong Zhou[4,5*], Yike Guo[3,8]

[1]Digital Medical Research Centre, School of Basic Medical Sciences, Fudan University, Shanghai, China
[2]Shanghai Key Laboratory of MICCAI, Shanghai, China
[3]Data Science Institute, Imperial College London, London, UK
[4]Shanghai Collaborative Innovation Centre of Endoscopy, Shanghai, China
[5]Endoscopy Centre and Endoscopy Research Institute, Zhongshan Hospital, Fudan University, Shanghai, China
[6]Academy for Engineering and Technology, Fudan University, Shanghai 200433, China
[7]School of Computer Science and Engineering, Nanjing University of Science and Technology, Jiangsu, China
[8]Department of Computer Science and Engineering, Hong Kong University of Science and Technology, Hong Kong, China

&These authors contributed equally
*Corresponding authors, li.quanlin@zs-hospital.sh.cn; zhou.pinghong@zs-hospital.sh.cn; shuowang@fudan.edu.cn.

**Abstract:** The development of artificial intelligence systems for colonoscopy analysis often necessitates expert-annotated image datasets. However, limitations in dataset size and diversity impede model performance and generalisation. Image-text colonoscopy records from routine clinical practice, comprising millions of images and text reports, serve as a valuable data source, though annotating them is labour-intensive. Here we leverage recent advancements in large language and vision models and propose EndoKED, a data mining paradigm for deep knowledge extraction and distillation. EndoKED automates the transformation of raw colonoscopy records into image datasets with pixel-level annotation. We validate EndoKED using multi-centre datasets of raw colonoscopy records (~1 million images), demonstrating its superior performance in training polyp detection and segmentation models. Furthermore, the EndoKED pre-trained vision backbone enables data-efficient and generalisable learning for optical biopsy, achieving expert-level performance in both retrospective and prospective validation.

Despite considerable improvements in prognosis over recent decades, colorectal cancer (CRC) remains the second leading cause of cancer mortality worldwide[1,2]. Colonoscopy plays a fundamental role in CRC screening programs that aim to reduce cancer incidence and mortality[3]. It is regarded the gold standard for detecting and removing precancerous colorectal polyps. Recent artificial intelligence (AI) systems for polyp detection and diagnosis have achieved promising progress towards improving adenoma detection rate[4] and enabling real-time determination of polyp histology during colonoscopy (i.e., optical biopsy)[5]. Several commercialised products are being translated into the clinical routine[6], while the AI performance remain to be further improved (e.g., generalisability and robustness to unseen data)[7].

To achieve clinical-grade performance, these deep learning models require large annotated datasets of lesion images which are expensive to collect prospectively[8]. Meanwhile, massive electronic colonoscopy records are generated and archived daily in the hospital information system[9], containing colonoscopy screenshots and free-text examination reports. Theses retrospective images represent a natural data source to address data scarcity but still require labour-intensive annotation. For computer vision tasks such as polyp detection and segmentation[10], experts are required to recognise polyp images and then draw bounding boxes or delineate polyp boundaries at the pixel level. Moreover, development of optical biopsy models further demands collecting benign and malignant polyps with confirmed histopathology[11]. The substantial annotation requirements constrain the size and diversity of datasets, preventing the full utilisation of historical unannotated data. It remains an open challenge to extract intrinsic supervision from raw colonoscopy records to train deep learning models with minimal annotation cost.

Self-supervised multimodal learning techniques such as Contrastive Language-Image Pre-training (CLIP)[12] enables representation learning from paired image-text data without reliance on manual annotations. Implicit representations of image and text are aligned so that semantically similar image-text pairs are close in the latent space. This approach succeeded in X-rays diagnosis (e.g., CheXzero[13], KAD[14]) by learning from radiology reports where each image is matched to a descriptive text. However, adapting CLIP to colonoscopy records is challenging as the image-text pairing is more complex. Dozens of images are captured from different anatomical locations during examination, while only a minority contain lesions of interest (Supplementary Figure S1). Extra pre-processing and alignment are required for semantic matching. Additionally, the colonoscopy report text tends to be more conclusive rather than descriptive, imposing difficulties for fine-grained representation learning. On the other hand, standard training of polyp detection and segmentation models requires explicit and detailed annotation of lesions on images (e.g., pixel-level segmentation mask) instead of the implicit representations.

Recent advancements in foundation models have revolutionized AI applications in medicine[15]. Large language models (LLMs) pre-trained on natural language have been utilised to analyse electronic health records and extract disease labels[16]. Meanwhile, large vision models (LVMs) like Segment Anything Model (SAM)[17] pre-trained on natural images, can delineate objects on medical images given appropriate prompts[18]. These motivate us to leverage recent advances in LLMs and LVMs to transform raw colonoscopy records into curated datasets with pixel-level annotation. In particular, we propose EndoKED, a knowledge extraction and distillation paradigm that connects LLMs and LVMs to mine intrinsic supervision within image-text colonoscopy records. We show that computer vision models trained with distilled knowledge from large-scale raw colonoscopy records achieve generalisable and robust performance on unseen datasets. Moreover, EndoKED provides a robust and flexible image encoder backbone enabling data- and time-efficient generalisation. This yields optical biopsy models comparable to clinical experts in multi-centre validation.

**Results**

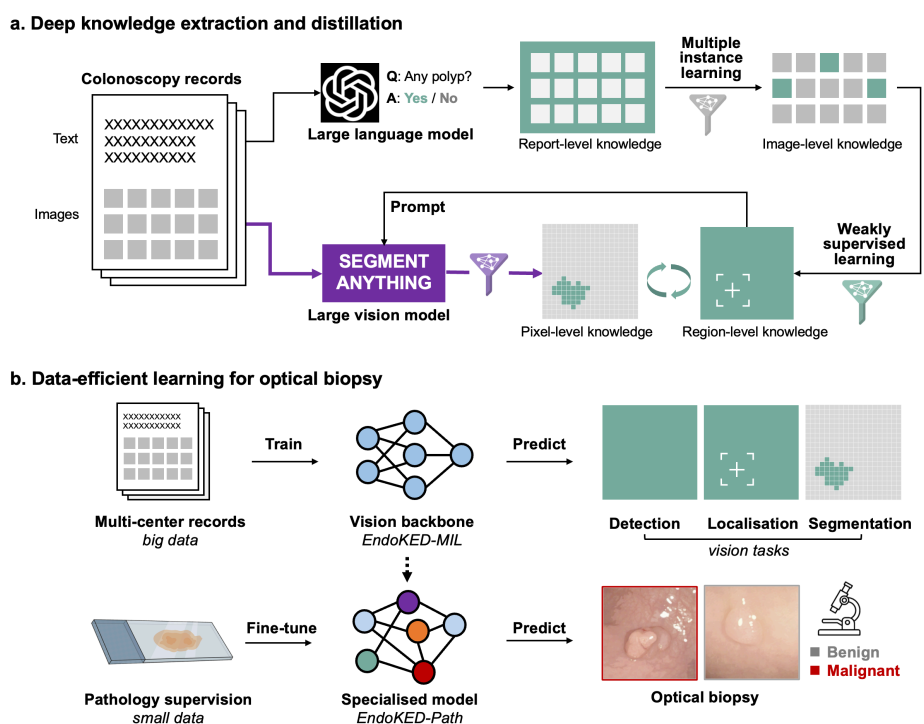

**Fig. 1 | Overview of the EndoKED design and applications to polyp diagnosis. a**, The intrinsic supervision from raw colonoscopy records is extracted leveraging large language and vision models. The report-level lesion label is firstly extracted from the free-text report by a large language model. The multiple instance learning (MIL) technique is used to propagate the report-level label to the image level. Then the region-level bounding box is obtained from class activation map (CAM). A large vision model, Segment Anything Model, takes the region-level bounding boxes as prompts to produce pixel-level lesion masks. Short description of this panel and of any symbols, marks and colours used. **b**, The image classification model for optical biopsy is developed in a data-efficient way – EndoKED pre-trained on multi-centre colonoscopy records then fine-tuned with limited pathology annotation.

**Overview of deep knowledge extraction and distillation**. As shown in Fig. 1a, the proposed data mining paradigm leverages synergies between a large language model and a large vision model, both general foundation models with non-medical pre-training. The LLM comprehends the raw text of the colonoscopy reports, while the LVM delineates corresponding objects from the images. This co-operation is enabled by a deep knowledge propagation and distillation design that automatically annotates the images. First, the LLM extracts report-level labels indicating lesion occurrence through question answering. These labels are then distilled into image-level labels by training a classification model to identify suspicious lesion images via multiple instance learning (MIL). Region-level bounding boxes are localised based on the class activation maps (CAM) and used to prompt the LVM to generate lesion masks. The distilled image- and pixel-level labels can then be utilised to train generic deep learning models for vision tasks such as lesion detection and segmentation. Furthermore, the pre-trained backbone can be fine-tuned in a data-efficient manner for clinical applications such as optical biopsy (Fig. 1b). We collected 14,177 colonoscopy reports with ~1 million endoscopic images to develop EndoKED, as detailed in Methods (Fig. 5). In this study, we focus on computer-aided detection and diagnosis of colorectal polyps, validated on independent retrospective and prospective datasets.

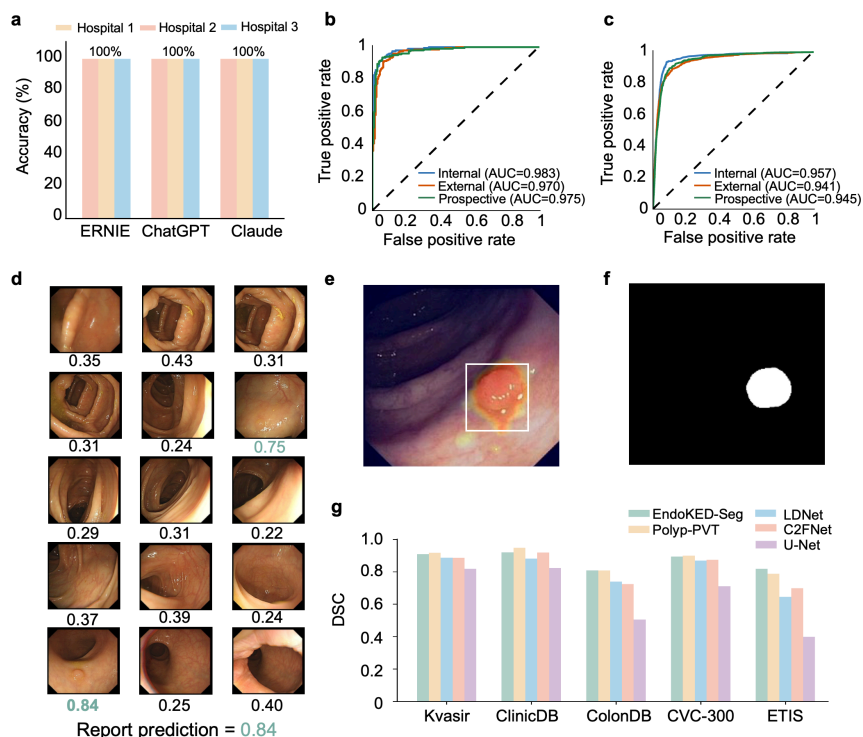

**Fig. 2 | Performance of the knowledge extraction and distillation paradigm for report abstraction, polyp detection and polyp segmentation. a**, Performance of large language models in extracting report-level label indicating polyp presence from free-text reports. **b**, Report-level performance of EndoKED-MIL in detecting polyp occurrence on three independent test datasets. **c**, Image-level performance of EndoKED-MIL in detecting polyp occurrence on three independent test datasets. **d**, Example polyp frame detection from dozens of screenshots within a colonoscopy record. The image-level prediction score is indicated below each image and the maximum is selected as the score for the report-level prediction. **e**, Example of region-level lesion detection within a polyp frame. **f**, Example of pixel-level polyp segmentation. **g**, Comparison of polyp segmentation performance on five public datasets between the proposed EndoKED trained on raw colonoscopy records and state-of-the-art segmentation models trained using expert annotation.

**Performance of LLMs in colonoscopy report abstraction.** The LLM was employed to comprehend free-text colonoscopy reports and extract report-level labels indicating polyp occurrence through designed question answering. We evaluated several state-of-the-art LLMs supporting Chinese language processing, including ChatGPT (GPT-3.5)[19], Claude[20] and ERNIE[21]. The LLMs were prompted to conclude whether any polyp was found in the raw text report. Performance was evaluated comparing to manual annotation of 300 colonoscopy reports randomly selected from the multi-centre datasets. As shown in Fig. 2a, all the LLMs achieved impressive high accuracy: ChatGPT Accuracy=100%; Claude Accuracy=100%; ERNIE Accuracy=100% in three independent hospitals. These results demonstrate that pre-trained LLMs can handle the task of report-level label extraction directly without further adaptation.

**Performance of multiple instance learning in lesion frame detection.** We developed a multiple instance learning method to propagate the knowledge of polyp occurrence from the report level to image level, as shown in Fig. 2d. A teacher network and a student network (EndoKED-MIL) for binary image classification were trained to predict report-level and image-level labels, respectively. The MIL method achieved high performance in both report and image-level polyp detection across three independent test sets. Report-level AUCs were 0.983, 0.970 and 0.975 (Fig. 2b) and image-level AUCs were 0.957, 0.941 and 0.945 (Fig. 2c) on the internal, external and prospective test set, respectively. Detailed classification metrics were provided in Supplementary Table 2.

**Table 1 | Performance comparison of EndoKED-SEG to state-of-the-art model trained with expert annotation on five public benchmark datasets.** Dice similarity coefficient (DSC) performance is shown, with Top 2 results in bold.

| Models | Kvasir | ClinicDB | ColonDB | CVC-300 | ETIS |
| --- | --- | --- | --- | --- | --- |
| U-Net[27] | 0.818 | 0.823 | 0.504 | 0.710 | 0.398 |
| U-Net++[28] | 0.821 | 0.794 | 0.482 | 0.707 | 0.401 |
| C2FNet[29] | 0.886 | 0.919 | 0.724 | 0.874 | 0.699 |
| DCRNet[30] | 0.886 | 0.896 | 0.704 | 0.856 | 0.556 |
| LDNet[31] | 0.887 | 0.881 | 0.740 | 0.869 | 0.645 |
| Polyp-PVT[32] | **0.917** | **0.948** | **0.808** | **0.900** | **0.787** |
| EndoKED-SEG | **0.908** | **0.920** | **0.809** | **0.893** | **0.818** |

**Performance of weakly supervised learning in lesion segmentation.** The image-level labels predicted by EndoKED-MIL were further distilled into pixel-level masks through SAM-guided weakly supervised semantic segmentation. As shown in Fig. 2e, the class activation maps of polyp detection were transformed to bounding boxes and then used as prompts to the SAM, enabling automatic generation of segmentation masks (Fig. 2f). Based on the distilled pixel-level annotation, we trained a polyp segmentation model (EndoKED-SEG) and evaluated its performance directly on five public benchmark datasets without any adaptation (Fig. 2g). EndoKED-SEG achieved competitive performance with averaged dice similarity coefficients (DSCs) of 0.908, 0.920, 0.809, 0.893, 0.818 on the Kvasir[22], CVC-ClinicDB[23], CVC-ColonDB[24], CVC-300[25] and ETIS[26] datasets, respectively (Table 1). This level of performance matches the state-of-the-art segmentation models[27–32] trained on datasets with expert annotation. Notably, EndoKED-SEG demonstrated superior generalisation and robustness on the most challenging ETIS dataset, with a 3% DSC improvement compared to the best counterpart.

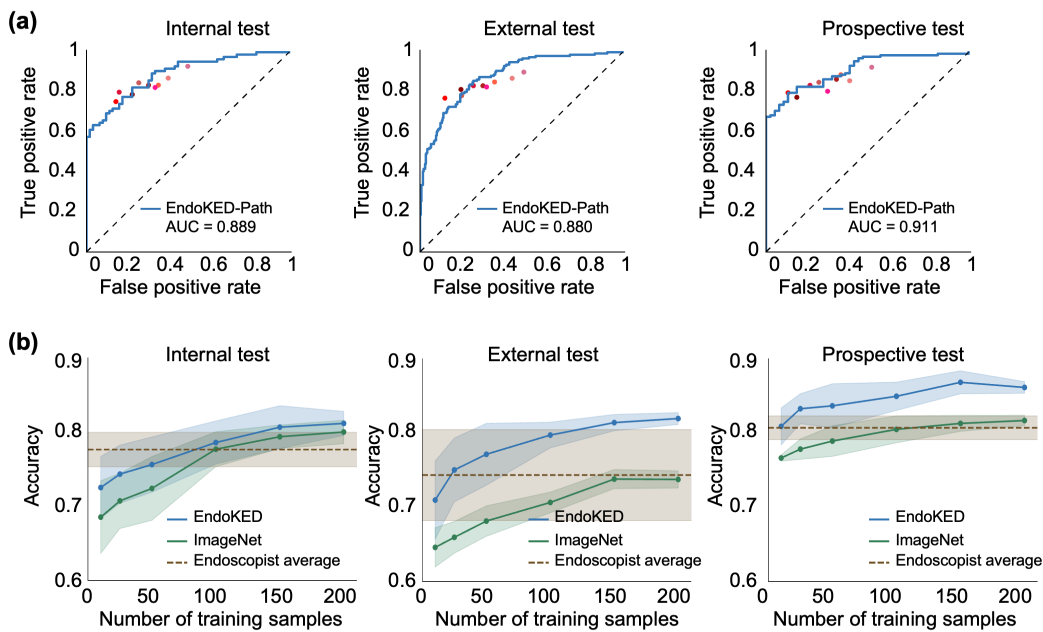

**Fig. 3 | Accuracy and data-efficiency of the EndoKED-Path optical biopsy model. a**, ROC curves of the EndoKED-Path model performance on the internal, external and prospective test sets compared to performance of nine endoscopists are plotted (dots). **b**, Data-efficient curves showing AUC improvement on the internal, external and prospective test sets as training set size increased during fine-tuning. Average and stand deviation of results from ten independently repeat sampling runs are plotted (solid line with shaded region). Average and stand deviation of nine endoscopist performance is shown for reference (dashed line with shaded region).

**Performance of transfer learning in optical biopsy**
The image encoder of EndoKED-MIL pre-trained on retrospective colonoscopy reports was adopted as the backbone for the prediction of histopathological malignancy through transfer learning. After fine-tuning on the pathology dataset, the resulting optical biopsy model EndoKED-Path achieved excellent AUCs of 0.889, 0.880 and 0.911 on the internal, external and prospective test sets, respectively (Fig. 3a). Performance was comparable to senior group of endoscopists, outperforming the average performance of junior endoscopists (Table 2). In comparison, the classification model with the same architecture of EndoKED-Path but pre-trained on ImageNet obtained lower AUCs of 0.833, 0.796 and 0.832 on the three test sets, respectively. Furthermore, EndoKD-Path demonstrated superior data-efficiency and generalisation ability – fine-tuning on only 10% of the data (20 samples) achieved similar performance to the ImageNet pre-trained model fine-tuned on the full dataset (200 samples) in the external and prospective validation (Fig. 3b).

**Table 2 | Performance comparison of EndoKED-Path optical biopsy model versus nine endoscopist and ImageNet pre-trained model.** AUC: area under curve of receiver operating curve, ACC: accuracy, SEN: sensitivity, SPE: specificity.

|  | Internal test | | | | External test | | | | Prospective test | | | |
| --- | --- | --- | --- | --- | --- | --- | --- | --- | --- | --- | --- | --- |
| Models | AUC | ACC | SEN | SPE | AUC | ACC | SEN | SPE | AUC | ACC | SEN | SPE |
| Senior #1 | - | 0.796 (0.794-0.798) | 0.753 (0.661-0.845) | 0.855 (0.767-0.943) | - | 0.838 (0.837-0.839) | 0.770 (0.705-0.835) | 0.877 (0.839-0.916) | - | 0.821 (0.820-0.823) | 0.797 (0.729-0.865) | 0.891 (0.801-0.981) |
| Senior #2 | - | 0.816 (0.814-0.818) | 0.800 (0.715-0.885) | 0.839 (0.747-0.930) | - | 0.772 (0.771-0.772) | 0.832 (0.775-0.890) | 0.736 (0.685-0.788) | - | 0.821 (0.820-0.823) | 0.835 (0.771-0.898) | 0.783 (0.663-0.902) |
| Senior #3 | - | 0.776 (0.773-0.778) | 0.835 (0.756-0.914) | 0.694 (0.579-0.808) | - | 0.742 (0.741-0.743) | 0.832 (0.775-0.890) | 0.690 (0.635-0.744) | - | 0.810 (0.808-0.812) | 0.865 (0.807-0.923) | 0.652 (0.515-0.790) |
| Senior #4 | - | 0.803 (0.801-0.805) | 0.847 (0.771-0.924) | 0.742 (0.633-0.851) | - | 0.790 (0.789-0.791) | 0.783 (0.719-0.846) | 0.794 (0.747-0.842) | - | 0.821 (0.820-0.823) | 0.850 (0.789-0.910) | 0.739 (0.612-0.866) |
| Senior #5 | - | 0.782 (0.780-0.785) | 0.788 (0.701-0.875) | 0.774 (0.670-0.878) | - | 0.804 (0.803-0.804) | 0.814 (0.754-0.874) | 0.798 (0.751-0.845) | - | 0.793 (0.792-0.795) | 0.774 (0.703-0.845) | 0.848 (0.744-0.952) |
| Senior average | - | 0.795 (0.794-0.795) | 0.805 (0.767-0.842) | 0.781 (0.735-0.827) | - | 0.789 (0.789-0.789) | 0.806 (0.779-0.834) | 0.779 (0.757-0.801) | - | 0.813 (0.813-0.814) | 0.824 (0.795-0.853) | 0.783 (0.729-0.836) |
| Junior #1 | - | 0.755 (0.753-0.758) | 0.835 (0.756-0.914) | 0.645 (0.526-0.764) | - | 0.712 (0.711-0.713) | 0.851 (0.796-0.906) | 0.632 (0.575-0.689) | - | 0.821 (0.820-0.823) | 0.887 (0.833-0.941) | 0.630 (0.491-0.770) |
| Junior #2 | - | 0.755 (0.753-0.758) | 0.871 (0.799-0.942) | 0.597 (0.475-0.719) | - | 0.664 (0.663-0.665) | 0.870 (0.818-0.922) | 0.545 (0.486-0.604) | - | 0.788 (0.786-0.790) | 0.857 (0.798-0.917) | 0.587 (0.445-0.729) |
| Junior #3 | - | 0.748 (0.746-0.751) | 0.929 (0.875-0.984) | 0.500 (0.376-0.624) | - | 0.639 (0.638-0.640) | 0.901 (0.854-0.947) | 0.487 (0.429-0.546) | - | 0.810 (0.808-0.812) | 0.925 (0.880-0.970) | 0.478 (0.334-0.623) |
| Junior #4 | - | 0.755 (0.753-0.758) | 0.824 (0.742-0.905) | 0.661 (0.543-0.779) | - | 0.728 (0.727-0.729) | 0.826 (0.768-0.885) | 0.671 (0.616-0.727) | - | 0.777 (0.775-0.778) | 0.805 (0.737-0.872) | 0.696 (0.563-0.829) |
| Junior average | - | 0.753 (0.753-0.754) | 0.865 (0.828-0.901) | 0.601 (0.540-0.662) | - | 0.686 (0.686-0.686) | 0.862 (0.835-0.888) | 0.584 (0.555-0.613) | - | 0.799 (0.798-0.799) | 0.868 (0.840-0.897) | 0.598 (0.527-0.669) |
| ImageNet Pre-trained | 0.833 (0.767-0.899) | 0.776 (0.773-0.778) | 0.741 (0.648-0.834) | 0.823 (0.727-0.918) | 0.796 (0.752-0.840) | 0.735 (0.734-0.736) | 0.795 (0.733-0.857) | 0.700 (0.646-0.754) | 0.832 (0.770-0.894) | 0.804 (0.803-0.806) | 0.842 (0.780-0.904) | 0.696 (0.563-0.829) |
| EndoKED-Path | 0.889 (0.839-0.940) | 0.776 (0.773-0.778) | 0.635 (0.533-0.738) | 0.968 (0.924-1.00) | 0.880 (0.847-0.912) | 0.781 (0.780-0.782) | 0.857 (0.803-0.911) | 0.736 (0.685-0.788) | 0.911 (0.870-0.952) | 0.821 (0.820-0.823) | 0.797 (0.729-0.825) | 0.891 (0.801-0.981) |

**Discussion**

This work demonstrates a pathway for mining intrinsic supervision from raw colonoscopy records through deep knowledge extraction and distillation. Progressive annotation of lesions was obtained on the report, image and pixel levels automatically. Deep learning models trained on 14,177 reports (containing ~1 million images) achieved impressive performance in computer vision tasks for colorectal polyp analysis. In particular, the EndoKED-SEG model accurately delineated polyps across benchmark datasets with an average DSC of 0.870, a level of performance previously reached only by models trained with expert-annotated datasets. To our best knowledge, this is the first time to achieve such pixel-level polyp segmentation performance based on unannotated colonoscopy records.

Previous attempts to train segmentation models with reduced annotation cost have also demonstrated performance enhancements. Semi-supervised methods have been used to train polyp segmentation models on datasets with fewer annotated samples. The knowledge of data distribution was exploited to propagate labels from annotated samples to unannotated ones. For example, Wu et al. proposed a semi-supervised model trained on datasets with 30% annotation that achieved DSCs of 0.89 on CVC-ClinicDB and 0.81 on Kvasir datasets[33], while performance dropped inevitably as the ratio of annotated data decreased. Other weakly supervised methods[33,34] also reduced annotation cost by replacing the pixel-level annotation with coarse annotation such as bounding boxes or scribbles. Nevertheless, these semi-supervised or weakly supervised methods still rely on the presence of certain expert annotations. The annotating cost remains considerable when scaling up to large datasets. In comparison, our proposed knowledge distillation approach does not require expert annotation and enables scaling up the training set size with millions of images.

The key innovation of EndoKED is the synergistic coupling of pre-trained LLMs and LVMs. Recent studies have validated the capability of LLMs to comprehend clinical text, which has also been confirmed in our study (Fig. 2a). Concurrently, LVMs such as SAM is being adapted for interactive medical image segmentation[18,35–37]. These observations encouraged us to devise an intuitive strategy to annotate colonoscopy reports – tasking LLMs with detecting lesion presence and SAM with delineating the regions of interest. However, each colonoscopy records encompasses dozens of endoscopic screenshots (~60 on average), imposing a discrepancy between the report-level labels generated by LLMs and the region-level prompts requisite for the LVM. To address this challenge, we developed a progressive knowledge distillation approach to propagate supervision across scales. The report-level label was transformed to image-level labels with multiple instance learning and further converted to pixel-level masks with weakly supervised segmentation techniques. Both utilised the teacher-student scheme that the teacher model is trained with coarse supervision while the student model yields fine-grained labels. EndoKED represents a generic framework where the LLMs and LVMs can be replaced with more advanced foundation models adapted for medicine. Moreover, the architectures of EndoKED-MIL and EndKED-SEG are flexible and compatible with different network design for image classification, lesion detection and segmentation. This enables both retrospective performance boosting of established computer-aided systems and further performance improvement alongside the latest progress in neural network design.

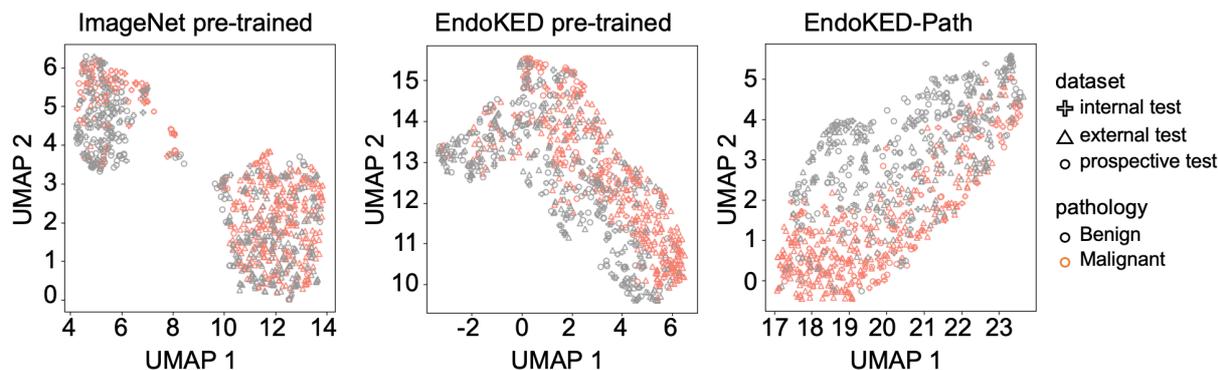

**Fig. 4 | Visualisation of the distribution of polyp images in the feature space.** Samples from different test sets are indicated with different symbols and the colour represents the malignance label according to histopathological classification.

A major advantage of models trained using EndoKED is its generalisability, attributed to exposure to substantial training data from real-world colonoscopy examination. This allows robust representation learning of various endoscopic patterns, empowering models to better recognize anomalies and segment abnormal findings. As shown in Table 1, EndoKED-SEG achieved excellent performance in polyp segmentation on five unseen public datasets, even surpassing models trained on the original datasets with expert annotation. In comparison, state-of-the-art segmentation models solely trained on the CVC-ClinicDB and Kvasir-SEG struggled on datasets with distribution shift such as ETIS, with performance dropping significantly. Similarly, EndoKED pre-trained optical biopsy models demonstrated consistent expert-level performance on independent internal, external and prospective test sets. In contrast, the ImageNet pre-trained version showed deteriorated performance on external and prospective datasets. This discrepancy is visualised in the feature spaces (Fig. 4). Samples from three test sets occupied isolated clusters in the ImageNet pre-trained feature space. The distribution became more compact after training on large-scale colonoscopy records, indicating improved features for polyp representation.

Furthermore, the generalisable polyp representation allowed for more efficient learning of the downstream clinical tasks. As depicted in Fig. 3b, the EndoKED pre-trained classification attained expert-level performance with substantially fewer fine-tuning samples. Specifically, this study achieved an averaged AUC of 0.890 in distinguishing malignant polyps across test datasets using only 230 cases of colonoscopy records with paired histopathological outcomes. The dataset required is remarkably smaller compared to similar studies[11,38,39], yet has accomplished comparable performance to senior endoscopists. The proposed transfer learning algorithm offers a promising solution to the challenge of annotating extensive datasets for the downstream clinical tasks, thereby advancing the application of artificial intelligence in the field of colonoscopy analysis.

In this study, we applied the EndoKED paradigm to obtain pixel-level annotation of polyps for the development of computer-aided polyp detection and diagnosis. It should be noted that the paradigm is readily extensible to analyse other lesions (e.g., bleeding) or endoscopic instruments (e.g., metal clips) in the report images, as long as their occurrence can be comprehended from the report text. Moreover, this paradigm could be potentially used for assisting knowledge discovery[40]. For instance, given a specific clinical concept (e.g., rare disease) or outcome (e.g., treatment response) recorded in the report, we could link the concept to relevant images and regions of interest for visualisation and hypothesis generation. The extracted endoscopic expertise highlights the potential to unlock and apply insights from massive amounts of unannotated data toward enabling more accurate and accessible precision diagnosis and care.

The study has the following limitations. Although the developed models were validated on benchmark image datasets, they have not been tested in real-world clinical settings. Additional validation in live colonoscopy procedures would be needed prior to clinical deployment. Potential biases may exist in the optical biopsy dataset. For example, diminutive rectal polyps (≤2 mm) judged by the endoscopist to be hyperplastic were typically not biopsied and therefore not included. This could limit model performance on small or histologically homogeneous polyps.

In conclusion, we have proposed and validated a data mining paradigm to extract pixel-level supervision from raw colonoscopy records, automated by the co-operation of large language and vision models. Deep learning models trained on retrospective colonoscopy records achieved impressive performance in polyp detection and segmentation. Furthermore, the pre-trained vision backbone enables data-efficient and generalisable learning of optical biopsy.

## Methods

**Multi-centre dataset of colonoscopy records.** The proposed EndoKED paradigm was developed and evaluated using a large-scale colonoscopy record dataset aggregated from four medical different centres, as shown in Figure 5a. Multiple colonoscopy records from the same patient were identified and only the most recent visit was included. The dataset comprises 14,177 colonoscopy records in total: 8,730 cases retrospectively collected between September to December 2022 along with 405 cases prospectively gathered in June 2023 from Zhongshan Hospital; 1,238 cases collected between March and April 2023 in Xiamen Branch of Zhongshan Hospital (Xiamen Hospital); 3,404 cases collected from Zhengzhou Central Hospital between March and April 2023 (Zhengzhou Hospital); 400 cases collected from No. 988 Hospital of Joint Logistics Support Force between January and March 2021 (No. 988 Hospital). For model development and validation, the retrospective reports from Zhongshan, Xiamen, and Zhengzhou hospitals were combined and randomly divided into a training set (13,372 reports) and an internal test set (400 reports). The external test set (400 reports) was constructed using cases from No. 988 Hospital, while the prospective test set (405 reports) consisted of cases from the prospective collection at Zhongshan Hospital. Each colonoscopy record contains dozens of endoscopic images and free-text report covering the examination findings, diagnostic results, recommendations, and colon preparation quality. For the internal, external, and prospective test sets, an experienced endoscopist (Z.Y.) manually reviewed the text report and images, labelling whether the report contained polyps and identifying images containing polyps. This expert annotation enabled quantitative evaluation of the model's ability to accurately detect polyps. Detailed information for each cohort was listed in Supplementary Table S1.

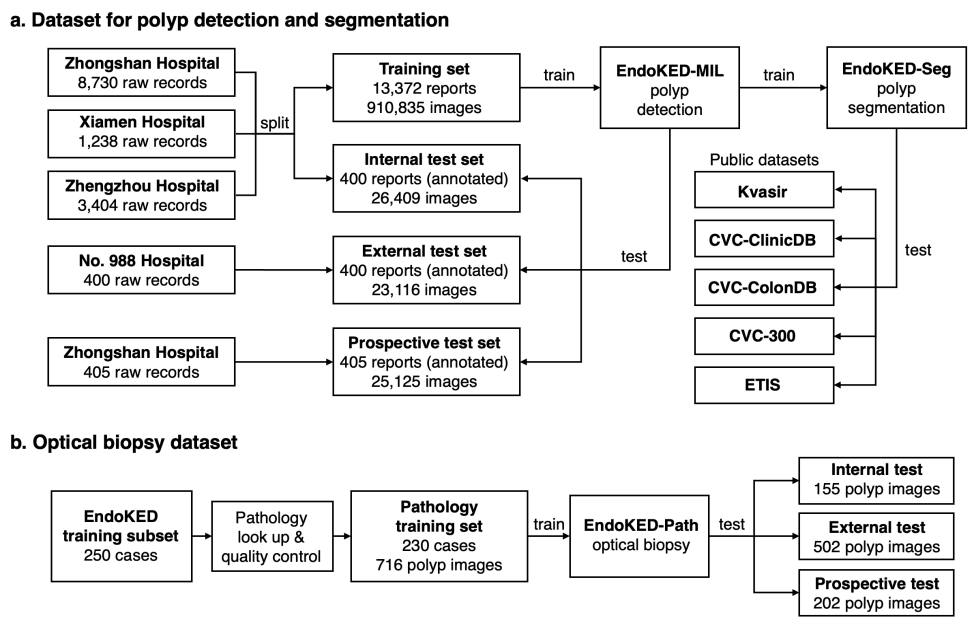

**Fig. 5 | Datasets for training and validating the EndoKED framework.** EndoKED is developed on retrospective datasets of colonoscopy records from three hospitals and validated on external and prospective datasets, as well as public datasets.

**Public datasets for polyp segmentation.** To evaluate the polyp segmentation model, five public datasets with expert annotations were utilised: Kvasir, CVC-ClinicDB, CVC-ColonDB, CVC-300 and ETIS (Supplementary Table S2). Each dataset contains endoscopic images of polyps along with corresponding segmentation masks delineating polyp boundaries. The segmentation performance was quantified by comparing the predicted polyp segmentation masks to the expert annotation using dice similarity coefficient.

The public datasets provide a rigorous benchmark comparing segmentation model's ability performance across different image sources, types of polyps, image qualities, and other variables.

**Optical biopsy dataset.** To develop and evaluate the optical biopsy model for predicting polyp malignancy, endoscopic images were extracted from the colonoscopy record dataset and looked up for histopathological outcome. The histopathologic classification followed the revised Vienna criteria for gastrointestinal neoplasia. Polyps categorised as Vienna 1 (negative for neoplasia) were considered benign, while polyps of category 3 (mucosal low-grade neoplasia), 4 (mucosal high-grade neoplasia), or 5 (submucosal invasion) were labelled as malignant that need further treatment. For model training, a subset of 250 colonoscopy records were randomly selected from the training set and reviewed. Inclusion criteria were patients with colorectal lesions (Vienna categories 1, 3, 4, and 5) that had confirmed pathology reports. Exclusion criteria included inflammatory bowel disease, familial adenomatous polyposis, and chemotherapy/radiation therapy. Poor quality images were also excluded. Three endoscopists (Z.Y., F.P., L.Q.) reviewed the training records, extracting 716 quality-controlled polyp images with malignancy labels. Similar extraction and quality control was conducted for the internal test (155 images), external test (502 images), and prospective test (202 images) sets, as shown in Figure 5b. To compare the diagnostic performance of the EndoKED-Path with endoscopists, we conducted an observational study on the test sets. Nine participating endoscopists, who were employees or trainees of Zhongshan hospital, were blinded to both the histopathological diagnosis and clinical information and asked to assign a binary malignant label of the polyps. Five endoscopists with more than five years of colonoscopy analysis experience were classified as the senior group, and four endoscopists with less than five years were classified as the junior group.

**Machine understanding of coloscopy report.** Three LLMs, i.e., ChatGPT, Claude, ERNIE, were employed to comprehend the free-text reports and extract report-level labels indicating the polyp presence through question answering. Specifically, each LLM is prompted to act as an endoscopist analysing the report text and concluding whether any polyp was found. The clear yes/no response from the LLM was encoded as 1 for the occurrence of polyp or 0 for the negative findings. The binary label served as the extracted report-level label for the following knowledge distillation. Detailed prompt design is provided in Supplementary Material.

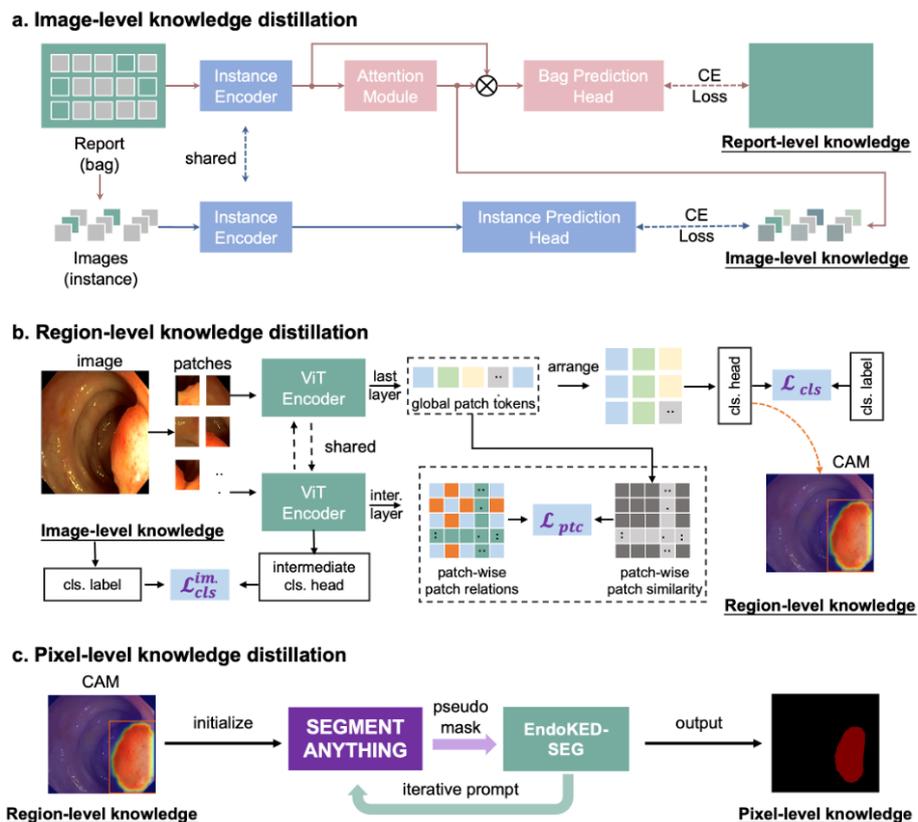

**Fig. 6 | Network design for cross-scale knowledge distillation.**

**Multiple instances learning for polyp frame extraction.** We developed a multiple instance learning framework EndoMIL to propagate the labels from the report level to the image level. As shown in Figure 6a, the MIL framework has a distillation structure with two heterogeneous branches, the teacher branch is a report-level classifier while the student branch is an image-level classifier. The teacher network is composed of an encoder and an attention-based pooling module followed by a bag prediction head. The student network (EndoKED-MIL) is composed of a shared encoder with teacher network but an instance prediction head. The teacher classifier is trained with report-level label and the student classifier is trained with image-level pseudo label obtained from teacher's attention score. Through distillation training, the student learns to classify images. In implementation, the encoder $f_t$ is implemented with ResNet18 and the pooling head $A_{t1}$ of teacher is implemented with attention-based deep multiple instance learning (ABMIL)[41]. The bag prediction head $\varphi_t$ and instance prediction head $\varphi_s$ is implemented with a fully-connected layer. To improve the stability of training, we further included another teacher network. The new teacher share encoder with original teacher and student, but the pooling head of the new teacher $A_{t2}$ is with design of dual-stream multiple instance learning network (DSMIL)[42]. The attention scores of the two teachers are merged with mean pooling to suppress the noise of pseudo labels:

$$\hat{Y}_i = Mean(A_{t1} \circ f_t(X_i), A_{t2} \circ f_t(X_i))$$

where the $X_i = \{x_{i,1}, x_{i,2} \dots x_{i,N}\}$ represents the report $i$ with $N$ images, and the $\hat{Y}_i = \{\hat{y}_{i,1}, \hat{y}_{i,2} \dots \hat{y}_{i,N}\}$ represents the pseudo labels of each image within the report $i$. The Cross Entropy (CE) is adopted to train the teachers and student:

$$\mathcal{L}_{t1} = CE(\varphi_s(A_{t1} \circ f_t(X_i) \cdot X_i^T), Y_i)$$
$$\mathcal{L}_{t2} = CE(\varphi_s(A_{t2} \circ f_t(X_i) \cdot X_i^T), Y_i)$$
$$\mathcal{L}_s = CE(\varphi_s \circ f_t(x_{i,j}), \hat{y}_{i,j})$$

where the $Y_i \in \mathbb{R}^1$ represents the ground truth report-level label. In inference, only student classifier is applied to obtain the image-level prediction:

$$\tilde{y}_{i,j} = \varphi_s \circ f_t(x_{i,j})$$

Then the report-level prediction is obtained by max-pooling:

$$\tilde{y}_i = \max_j \tilde{y}_{i,j}$$

Details of the network architecture and training can be found in Supplementary Material.

**Weakly supervised learning for polyp localisation and segmentation.** We developed a SAM-guided weakly supervised semantic segmentation (WSSS) technique to segment polyps given image-level labels. This is realised by the region-level localisation through class activation map (CAM) followed by pixel-level annotation empowered by Segment Anything Model (SAM). CAM is a typical method to distil region-level (or coarse pixel-level) knowledge from image-level classification label but facing notorious challenges of incomplete and inaccurate localisation of objects. In this study, we adopted a vision transformer (ViT)-based framework[43] to address these problems. To alleviate the incompleteness, ViT is used as the classification encoder whose inherent long-range dependency qualifies the CAM to cover the whole object. To generate more accurate localisation clues, patch-wise relations are captured from the intermediate layers of ViT to avoid the incorrect activation of CAM from the last layers. As shown in Figure 6b, with the image $X$ and class label $Y$, we split the image into $N \times N$ patches $x_i$. Then the encoders extract a patch token $t_i$ for each patch. After arranging the patch tokens $T \in \mathbb{R}^{N^2 \times C}$ into spatial feature map $F \in \mathbb{R}^{N \times N \times C}$, spatial feature map is operated with a global max pooling and sent to a classification layer to calculate the classification loss $\mathcal{L}_{cls}$. Multi-label soft margin loss is used to supervise the classification, which is formulated as:

$$\mathcal{L}_{cls} = \frac{1}{K}\sum_{k=1}^{K}(Y^k log(p_{cls}^k) + (1-Y^k)log(1-p_{cls}^k))$$

where $Y^k$ is the class label for category $k$, $p_{cls}^k$ is the class probability vector from the last layer. Then the CAM is generated by weighting and summing the feature maps with the weights $W \in \mathbb{R}^{K \times C}$, where $K$ is the number of categories. The process can be formulated as:

$$CAM_k = \frac{relu(\sum_i W_{k,i} F_{:,i})}{max(relu(\sum_i W_{k,i} F_{:,i}))}$$

where $relu(\cdot)$ and $max(\cdot)$ are the activation function and the max normalization. To avoid the inaccurate activation of CAM caused by the over-smoothness from ViT, more fine-grained and diverse relations between patches are extracted from the intermediate layers of the encoder, which is utilized to supervise the relations from the last layer. The similarity matrix of patch tokens is extracted by computing the cosine similarity among all patch tokens from the last layer and the intermediate layer, respectively. With the similarity matrix $M_s \in \mathbb{R}^{N^2 \times N^2}$ and $M_t \in \mathbb{R}^{N^2 \times N^2}$ from the last layer and the intermediate layer, InfoNCE-basd patch token contrast loss $\mathcal{L}_{ptc}$ is leveraged to alleviate the mis-activation. The process can be formulated as:

$$\mathcal{L}_{ptc} = \frac{1}{N^+}\sum_{Y_i=Y_j}(1-CosSim(F_i, F_j)) + \frac{1}{N^-}\sum_{Y_i \neq Y_j}CosSim(F_i, F_j)$$

where $CosSim(\cdot)$ calculates the cosine similarity and $N^+/N^-$ is the number of positive/negative pairs determined by the relations from $M_t$. In addition, to guarantee the correctness of relations from the intermediate layer, we employ an auxiliary classification loss $\mathcal{L}_{cls}^{im}$, which is in form of multi-label soft margin loss as well. With the region-level knowledge from CAM, the bounding box prompt is sent to the large vision model (SAM) to distil more fine-granularity pixel-level knowledge. The knowledge distillation follows such pipeline as shown in Figure 6c. With the initialised prompt from CAM, pseudo masks are generated from SAM. Then the pseudo masks are sent to a polyp segmentation model to train the decoder and output segmentation mask as pixel-level annotation. To refine the segmentation, the bounding box is extracted from the prediction and used as the prompt of SAM to produce more accurate pseudo mask. Such process is iteratively conducted until the final pixel-level annotation is reliably distilled from the large vision model to the segmentation model. Details of the network architecture and training can be found in Supplementary Material.

**Transfer learning for optical biopsy.** To transfer knowledge from the polyp detection task to the adenoma prediction task, the prediction head in EndoKED-MIL is replaced with a new light weight prediction head composed of two fully-connected layers with random initialised weights. Then the resulting EndoKED-Path network is fine-tuned on the optical biopsy training set and evaluated on the three test sets.

**Reporting Summary**.
Further information on research design is available in the Nature Research Reporting Summary linked to this article.

**Data availability**
The main data supporting the findings of this study are available within the Article and its Supplementary Information. The raw data generated in this study are available from the corresponding author upon reasonable request.

**Code availability**
The codes and trained models of this study are publicly available at
https://github.com/shuowang26/EndoKED.

**Acknowledgements**
This study was supported by the National Natural Science Foundation of China (82203193), Shanghai Sailing Programs of Shanghai Municipal Science and Technology Committee (22YF1409300), International Science and Technology Cooperation Program under the 2023 Shanghai Action Plan for Science (23410710400).

**Author contributions**
S.W., Y.Z., Q.L, P.Z and Y.G. conceptualised the study. S.W. and Y.Z. designed the study. S.W., Y.Z., X.L., Z.Y., P.F. and Y.Z.  performed data analysis and interpretation. S.W., Y.Z., X.L. and Z.Y. drafted the manuscript. M.W., Z.S., Q.L. and P.Z. supervised the study and revised the manuscript.

**Inclusion & Ethics**
This study was approved by the institutional review board of each participating hospital. Written informed consent was waived for the retrospective and external cohorts, and was obtained for the prospective cohort. Colonoscopy records were acquired from the information system of each hospital.

**Competing interests**
The authors declare no competing interests.

**Additional information**
Supplementary figures and tables are provided in the Supplementary Material.

# Supplementary Materials

# Knowledge Extraction and Distillation from Large-Scale Image-Text Colonoscopy Records Leveraging Large Language and Vision Models

## Table of Contents

## 1. Collection of colonoscopy records

In Chinese hospitals, patients often receive a printed examination report after colonoscopy (Figure S1a). The printed report includes free-text description of the procedure and findings, as well as some optional images. It is noted that not every hospital includes images in the printout but archives them in electronic health record system only. In this study, we collected the full report text and all images captured during colonoscopy from the hospital information system (Figure S1b). This provided more complete data for developing and evaluating our proposed EndoKED approach.

The colonoscopy records were collected from four hospitals and split into different cohorts including the training, internal test, external test and prospective test sets, as described in the Method Section. The clinical characteristics of each cohort are listed in Table S1. The training set was used to develop the EndoKED model. The internal test set was from the same distribution as the training set. The external test set was from different hospitals to evaluate generalisability. The prospective test set was collected after model development for independent validation. This cohort splitting strategy enabled comprehensive evaluation of EndoKED performance.

## 2. Prompts for extracting report-level label

We leveraged the Chat completions API from ChatGPT to extract the report-level label from the free-text description. The model we select for large-scale distillation is GPT-3.5-Turbo. The main input for the model is the message parameter. Messages has a role (either "system", "user"or "assistant") and content. The content is the prompt for the model to conduct an ordered response. For the content of system message, we asked the LLM to have a personality of a professional endoscopist, i.e., {"You are a professional endoscopist"}. For the content of user role, we provided the clinic report for the assistant to respond to:

{"Based on the following text provided, determine whether the content refers to polyps. Flat and raised lesions are also a type of polyps. If there are polyps, answer 1; if there are no polyps, answer 0. You only need to answer 1 or 0, no explanation is required. Here is the text: ***report***"}.

The message for assistant includes the chat history from the LLM so that the question and answering can be proceeded. More details can be found in our codes.

## 3. EndoKED-MIL for polyp detection

In the weakly-supervised distillation framework, the collection all images corresponding to the colonoscopy report is considered as a 'bag' and each image is an 'instance'. For the instance encoder, we adopted the encoder part of ResNet18 and removed its last classification layer. Both the bag prediction head and the instance prediction head were implemented using a two-layer fully connected network. In the framework, the attention modules for the two teacher branches used attention-based deep multiple instance learning (ABMIL)[1] and dual-stream multiple instance learning network (DSMIL)[2], respectively. The Attention module $A_{t1}$ of ABMIL consists of a two-layer fully connected network. The pooling method for ABMIL is:

$$a_i = A_{t1}(f_t(x_i))$$

$$F = \sum_i \frac{\exp(a_i)}{\sum_j \exp(a_j)} \cdot f_t(x_i)$$

where $x_i$ is the $i$-th instance in the bag, $a_i$ is the corresponding attention score, and $F$ is the aggregated bag feature. The DSMIL employs a non-local attention mechanism. First, it trains an instance-level classifier to obtain the critical instance:

$$c = \max_i g(f_t(x_i))$$

$$x_{critical} = \underset{x_i}{\operatorname{argmax}}\, g(f_t(x_i))$$

Then, the non-local attention scores are calculated based on the distance between other instances and the critical instance:

$$q_i = f_q(f_t(x_i))$$

$$v_i = f_v(f_t(x_i))$$

$$a_i = \frac{\exp(\langle q_i, q_{critical}\rangle)}{\sum_j \exp(\langle q_j, q_{critical}\rangle)}$$

$$F = \sum_i a_i \cdot v_i$$

where $f_q$ and $f_v$ are transforming networks, both implemented with a single fully connected layer, and $a_i$ is the non-local Attention score. We compute the mean attention scores from both ABMIL and DSMIL to create the pseudo instance label and subsequently train the student network.

During training, we alternately train the teacher branch and the student branch, training the teacher branch for one epoch and then the student branch for one epoch, and continue in this manner. We used the SGD optimizer with a learning rate of 0.001. During training, we employed random data augmentations, including random resized crop, random horizontal flip, random vertical flip, random affine transformation, and random color jitter. The classification performance evaluated on the internal, external and prospective test sets, respectively (Table S2).

## 4. EndoKED-SEG for polyp segmentation

To extract the region-level knowledge, we adopted ViT-B/16 as the encoder in our framework. The input image is first resized to $448 \times 448$ and then split into patches with the size of $16 \times 16$. The encoder has 12 layers in total and 10th layer are preferred to generate the similarity matrix as the intermediate layer. We train the classification model with AdamW optimizer and the learning rate linearly increases to 6e-5 with a polynomial scheduler.

To generate reliable pseudo masks, we gradually refine the masks with Segment Anything Model (SAM). We generate bounding box from the initial class activation map (CAM) and send it as prompt to SAM. SAM leverage the bounding box to segment polyps masks. With this mask, we additionally train a segmentation model, Polyp-PVT[3], then all the predicted masks from the segmentation model are used as prompt for SAM for further refinement. With these updated masks, the segmentation model is iteratively optimised and becomes more reliable. The architecture of the segmentation model is shown in Figure S2. It is comprised of a pyramid-ViT-based encoder (a), a cascade fusion module (b), a camouflage identification module, and a similarity aggregation module (d). Cascade fusion module plays an important role in fusing the high-level feature. Camouflage identification module filters out the low-level information. Similarity aggregation module efficiently integrate the information from the high and low-level features. More details for the settings please refer to polyp-PVT.

Based on the massive reports and images, the segmentation model demonstrates an impressive generalization ability on five public out-of-domain datasets, i.e., CVC-ClinicDB[4], Kvasir-SEG[5], ETIS[6], CVC-ColonDB[7], and CVC-300[8]. Following the common experimental setups[3], the training set from CVC-ClinicDB and Kvasir-SEG are not included and we evaluate our model only in the testing set for a fair comparison. The detailed description for the datasets is reported in Table S3.

## 5. EndoKED-Path for optical biopsy

To verify the generalizability and transferability of the encoder trained by EndoKED-MIL, we constructed several few-shot datasets with pathological annotations. Next, we contrasted the prediction results from ResNet18 pretrained using EndoKED, ResNet18 pretrained on ImageNet, and the diagnostic assessments of endoscopists. For the data-efficiency experiment, we randomly sampled different numbers of samples from the training set to construct various few-shot training sets. Considering the randomness of the selected training samples greatly affects few-shot experiments, for each few-shot training set, we randomly constructed it five times, and reported the average and variance of the network's accuracy on the test set after these five trainings. We conducted experiments under settings of 10-shot, 20-shot, 50-shot, 100-shot, 150-shot, and 200-shot.

    For both EndoKED pre-trained ResNet18 and ImageNet pre-trained ResNet18, we adopted an identical fine-tuning strategy. We replaced the prediction head of pretrained ResNet18 with two fully connected layers, and then fine-tuned all parameters of the entire network. During training, we employed data augmentation strategies, including random resized crop, random horizontal flip, random vertical flip, random affine transformation, and random color jitter. During testing, we reported the accuracy, sensitivity, and specificity of the models at the operation point with maximum Youden index.

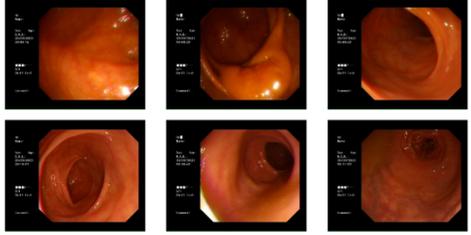
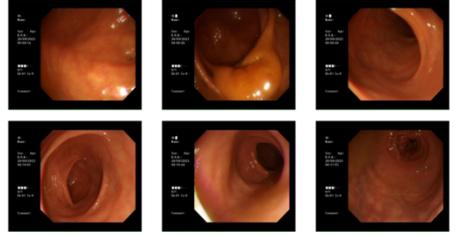
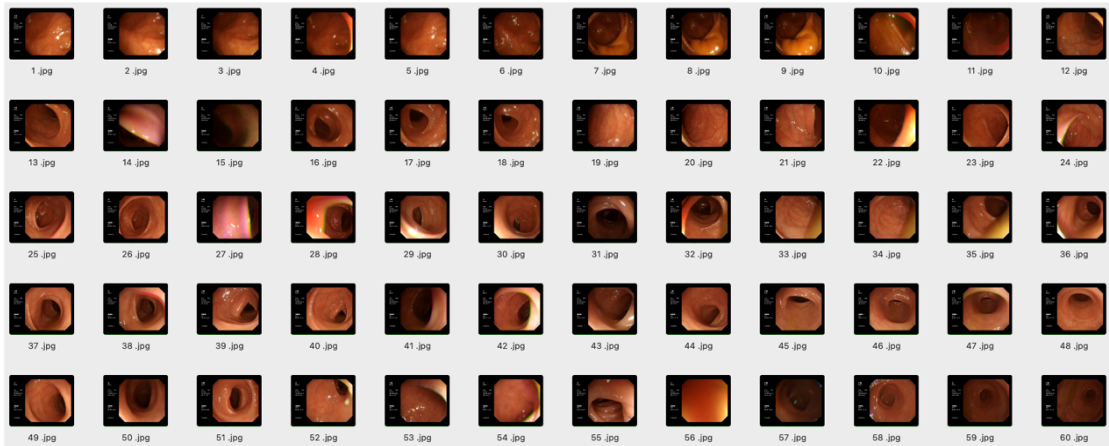

**Supplementary Figure S1. Examples of image-text colonoscopy records. a)** An example of printed examination report in Chinese including text description (indicated in the red box with dash lines) and several images. Translated version in English is on the right for reference. **b)** All screenshot images corresponding to the printed report.

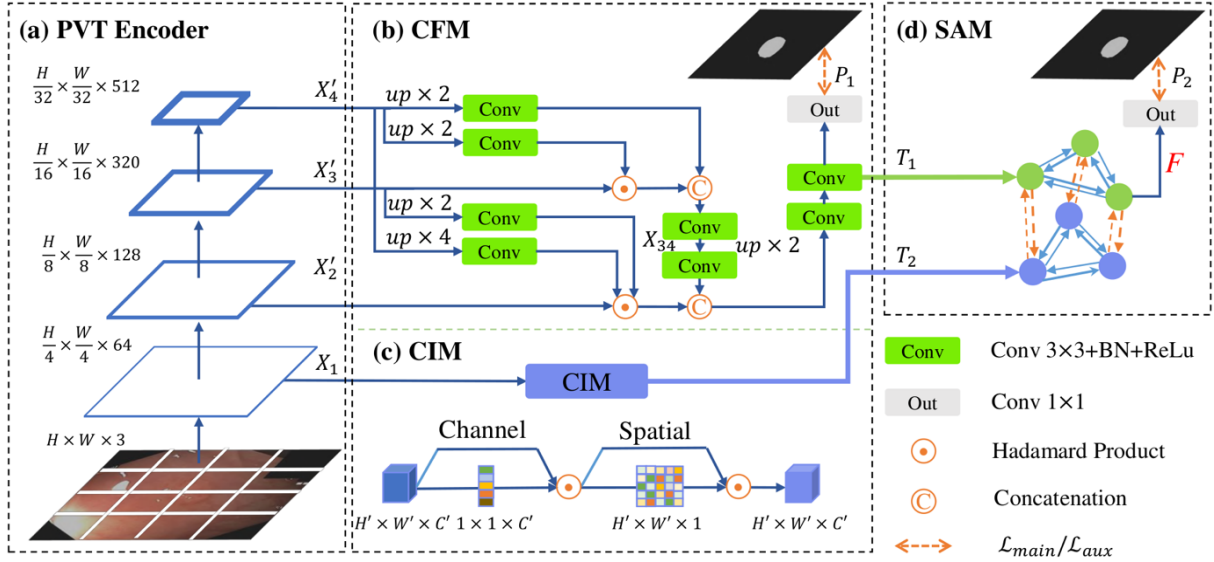

**Supplementary Figure S2.** Network architecture of the segmentation model Polyp-PVT[3].

**Supplementary Table S1. Clinical characteristics of different cohorts.**

|  | Training set | Internal test set | External test set | Prospective test set |
|---|---|---|---|---|
| **Number of colonoscopy records** | 13,372 | 400 | 400 | 405 |
| **Number of images** | 910,835 | 26,409 | 23,116 | 25,125 |
| **Median age** | 61 | 62 | 58 | 62 |
| **Male sex (%)** | 6,645 (49.7%) | 204 (51.0%) | 228 (57.0%) | 210 (51.9%) |
| **Number of reports with polyps*** | - | 200 | 200 | 205 |
| **Number of images with polyp*** | - | 451 | 661 | 472 |
| **Histopathology outcome** | - | 200 | 200 | 205 |

*Colonoscopy records in internal, external and prospective test sets were manually reviewed to identify polyp occurrence and corresponding polyp images.

**Supplementary Table S2. Report-level and image-level classification performance in polyp detection on three independent test sets.** AUC: area under curve of receiver operating curve, ACC: accuracy, SEN: sensitivity, SPE: specificity.

| | Metric | Internal test | External test | Prospective test |
|---|---|---|---|---|
| **Report-level** | AUC | 0.983 (0.974-0.993) | 0.970 (0.955-0985) | 0.975 (0.962-0.988) |
| | ACC | 0.938 (0.937-0.938) | 0.922 (0.922-0.923) | 0.938 (0.938-0.939) |
| | SEN | 0.935 (0.901-0.969) | 0.910 (0.870-0.950) | 0.937 (0.903-0.970) |
| | SPE | 0.940 (0.907-0.973) | 0.935 (0.900-0.969) | 0.940 (0.907-0.973) |
| **Image-level** | AUC | 0.957 (0.949-0.965) | 0.941 (0.932-0.950) | 0.945 (0.937-0.954) |
| | ACC | 0.914 (0.914-0.914) | 0.876 (0.876-0.876) | 0.896 (0.896-0.896) |
| | SEN | 0.938 (0.916-0.910) | 0.896 (0.872-0.919) | 0.898 (0.871-0.926) |
| | SPEC | 0.914 (0.910-0.917) | 0.875 (0.871-0.880) | 0.896 (0.893-0.900) |

**Supplementary Table S3. Description of five public datasets for polyp segmentation.**

| Dataset | Year | Resolution | Training | Testing | Total |
|---|---|---|---|---|---|
| CVC-ClinincDB[4] | 2015 | 384×384 | 550 | 62 | 612 |
| Kvasir-SEG[5] | 2020 | 332×487~1920×1072 | 900 | 100 | 1000 |
| ETIS[6] | 2014 | 1225×966 | N/A | 196 | 196 |
| CVC-ColonDB[7] | 2016 | 574×500 | N/A | 380 | 380 |
| CVC-300[8] | 2017 | 574×500 | N/A | 60 | 60 |